\documentclass[runningheads]{llncs}
\usepackage[T1]{fontenc}
\usepackage{graphicx}
\usepackage{tabularx}
\usepackage{booktabs}
\usepackage{array}
\usepackage{multirow}
\usepackage[normalem]{ulem}
\usepackage{siunitx}
\usepackage{amsmath}  
\begin{document}
\title{IMPA-HGAE:Intra-Meta-Path Augmented Heterogeneous Graph Autoencoder}
%
%

\author{Di Lin\inst{1} \and Wanjing Ren\inst{1} \and Xuanbin Li\inst{2} \and Rui Zhang\inst{3}}

\institute{
College of Software, Jilin University \email{lindi5522@mails.jlu.edu.cn} \\
College of Software, Jilin University \email{renwj5522@mails.jlu.edu.cn} \\
College of Software, Jilin University \email{lixb5522@mails.jlu.edu.cn} \\
College of Computer Science and Technology, Jilin University \email{rui@jlu.edu.cn}
}

\maketitle              
\begin{abstract}
Self-supervised learning (SSL) methods have been increasingly applied to diverse downstream tasks due to their superior generalization capabilities and low annotation costs. However, most existing heterogeneous graph SSL models convert heterogeneous graphs into homogeneous ones via meta-paths for training, which only leverage information from nodes at both ends of meta-paths while underutilizing the heterogeneous node information along the meta-paths. To address this limitation, this paper proposes a novel framework named IMPA-HGAE to enhance target node embeddings by fully exploiting internal node information along meta-paths. Experimental results validate that IMPA-HGAE achieves superior performance on heterogeneous datasets. Furthermore, this paper introduce innovative masking strategies to strengthen the representational capacity of generative SSL models on heterogeneous graph data. Additionally, this paper discuss the interpretability of the proposed method and potential future directions for generative self-supervised learning in heterogeneous graphs. This work provides insights into leveraging meta-path-guided structural semantics for robust representation learning in complex graph scenarios.

\end{abstract}
\section{Introduction}

In the real world, graph structures often do not strictly obey to the presumption of homogeneous association, but instead exhibit an opposite characteristic—heterogeneous characteristic that linked nodes have dissimilar features and different class labels. \cite{zheng2022graph} Due to the powerful capability of heterogeneous graphs in modeling complex systems, an increasing number of researchers have started to focus their attention on the study of GNN with heterogeneous characteristics. Most Heterogeneous Graph Neural Networks (HGNNs) \cite{hu2020heterogeneous} \cite{tian2021recipe} \cite{zhang2019heterogeneous}  rely on labeled data during training, following the paradigms of supervised or semi-supervised learning. The main challenge with this approach is that acquiring labeled data can be expensive or insufficient. \cite{wang2022graphfl} Therefore, leveraging self-supervised learning to explore data's inherent characteristics for enhanced feature extraction is a promising solution.

\begin{figure}[htbp]
    \centering
    \includegraphics[width=0.8\textwidth]{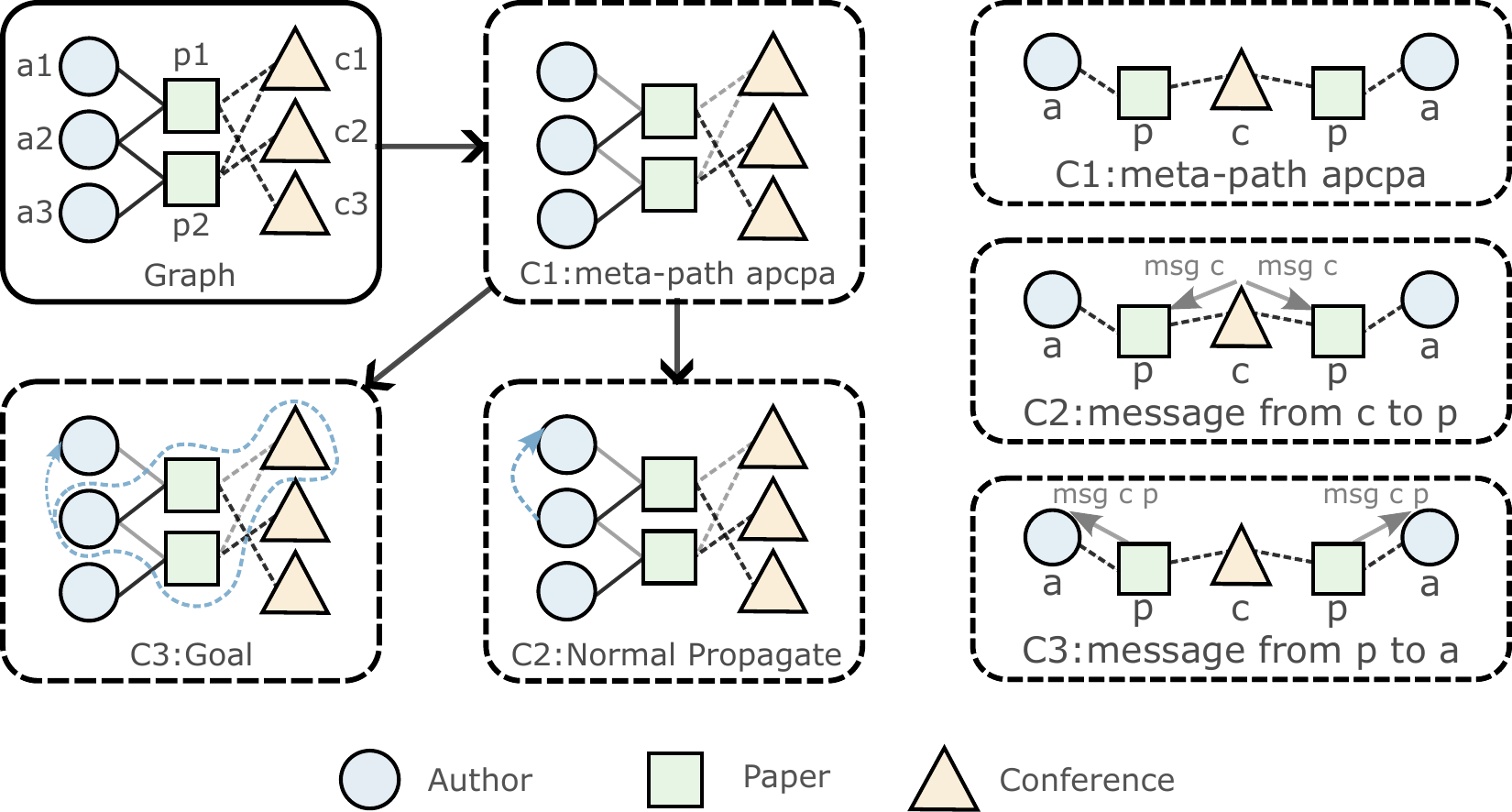}
    \caption{Comparative Analysis: Real-World Challenges in Heterogeneous Graphs (Left) vs. IMPA-HGAE's Innovative Solutions (Right) \\ (Left) Conventional approach: (C1) APCPA metapath illustration, (C2) Generic neighborhood aggregation, (C3) Intermediate node-agnostic propagation. \\ (Right) IMPA-HGAE framework: (C1) Annotated metapath, (C2) Author-to-paper relation learning, (C3) Paper-to-affiliation attribute transfer.}
    \label{fig:example1}
\end{figure}

Self-supervised graph auto-encoder GraphMAE \cite{hou2022graphmae} proposes a method for masked reconstruction of node features in graphs. By extending the core idea of GraphMAE to heterogeneous graphs, HGMAE \cite{tian2023heterogeneous} has been derived.

This series of masked auto-encoder models demonstrates an evolutionary progression in architecture design, but one thing they have in common is that when processing datasets with longer meta-paths containing richer node information, they all consider only the node information at both ends of the meta-paths. GraphMAE is designed for homogeneous datasets, i.e., single-category, multi-label data. Although HGMAE targets heterogeneous datasets i.e.,multi-category, multi-label data, both approaches overlook the heterogeneous semantics and the feature information of intermediate nodes in meta-paths. For instance, in the DBLP dataset, predefined meta-paths (e.g., Author-Paper-Author) transform the adjacency matrix into an A-P-A format, but the intermediate node (P) features are ignored.

Masked autoencoder training for heterogeneous graphs should preserve the rich semantics and structure from diverse nodes and complex interactions. Traditional methods primarily focus on structural information while neglecting both heterogeneous semantics and the feature information of intermediate nodes, resulting in suboptimal performance.

\textbf{Contributions.} Considering the above requirements, this paper proposes a heterogeneous graph autoencoder called IMPA-HGAE. The contributions are as follows:
\begin{enumerate}
  \item This paper proposes IMPA-HGAE, a novel approach for node aggregation along meta-paths in heterogeneous graphs. In addition, this paper discusses potential future research directions for generative self-supervised modeling.
  \item This paper presents a strategy for enhancing the representation of a target node by exploiting neighboring nodes on a meta-path in a graph. This approach effectively utilizes both the structural information of heterogeneous graphs and the rich semantic information carried by nodes within meta-paths.
  \item Extensive experimental results demonstrate that IMPA-HGAE outperforms various existing baseline models across different datasets. Additionally, ablation studies provide interpretability analysis for the generative self-supervised model by examining the impact of different masking strategies on noise introduction.
\end{enumerate}

\section{Related Work}

The work presented here is closely related to heterogeneous graph neural networks and graph self-supervised learning. In this section, a brief overview of these two aspects is provided.

\textbf{Heterogeneous Information Network.} For a heterogeneous graph \( \mathcal{G}=\left(\mathcal{V}, \mathcal{E}, \mathcal{A}, \mathcal{X}, {\mathcal{T}_{v}}, \mathcal{T}_{\varepsilon}, \Phi\right) \), \( \mathcal{V} \) represents the set of nodes, \( \mathcal{E} \) represents the set of edges, represents the set of nodes, \( \mathcal{A} \) represents the adjacency matrix, \( \mathcal{X} \) represents the feature matrix, \( \mathcal{T}_{v} \) represents the types of nodes, \( \mathcal{T_{\varepsilon}} \) represents the types of edges and \( \Phi \) represents the set of meta-paths.  For example, \( \mathcal{T}_{\nu_1} \xrightarrow{\mathcal{T}_{\epsilon_1}} \mathcal{T}_{\nu_2} \xrightarrow{\mathcal{T}_{\epsilon_2}} \cdots \xrightarrow{\mathcal{T}_{\epsilon_l}} \mathcal{T}_{\nu_{l+1}} \) is a meta-path of length \( l \).  \(\mathcal{A}^{\phi} \) ,\(  {\phi} \in {\Phi} \)  is the adjacency matrix based on the metapath  \( \phi\).

Recent research has increasingly focused on heterogeneous graph neural networks, primarily employing two key strategies: non-local neighbor extension and GNN architecture refinement \cite{zheng2022graph}. Non-local neighbor extension methods aggregate messages from multi-hop neighbors through various approaches. For instance, MixHop \cite{abu2019mixhop}, H2GCN \cite{zhu2020beyond}, and UGCN \cite{wang2020motion} directly model multi-hop interactions. GNN structural refinement enhances the representation of homogeneous and heterogeneous information through diverse strategies. FAGCN \cite{bo2021beyond} and ACM \cite{luan2021heterophily} refine message aggregation, H2GCN separates ego and neighbor representations, and GPR-GNN \cite{chien2020adaptive} optimizes inter-layer information flow. However, these methods rely on labeled nodes and cannot learn self-supervised embeddings directly from data. A key challenge remains: effectively leveraging heterogeneous graph structures for self-supervised representation learning.

\textbf{Self-supervised Learning on Graphs.} Graph self-supervised learning has been widely applied to learning representations from unlabeled graphs. Existing learning methods can be broadly categorized into contrastive learning and generative learning \cite{liu2021self}.  While contrastive learning methods have been the dominant approach in the field of graph self-supervised learning over the past few years, the recent emergence of Graph Masked Autoencoders (GraphMAE) has renewed interest in generative methods. Contrastive learning has shown significant advantages in the field of graph neural networks, with typical representatives including HeCo \cite{wang2021self}, HDGI \cite{ren2019heterogeneous} , and DMGI \cite{cheng2023wiener}. However, the success of these methods largely depends on the careful design of negative sampling, architecture, data augmentation, and other factors \cite{you2021graph}. The goal of generative graph self-supervised learning is to use the input graph as self-supervision to recover missing parts of the input data. For example, GraphMAE aims at feature reconstruction and introduces a masked graph autoencoder for self-supervised graph training. HGMAE further develops this idea by proposing two masking techniques: meta-path masking and adaptive attribute masking with dynamic mask rate, along with three training strategies: edge reconstruction based on meta-paths, target attribute restoration, and positional feature prediction. Additionally, within the research field of Graph Autoencoders (GAE) \cite{kipf2016variational}, there exist similar models such as WDGN \cite{cheng2023wiener}, SeeGera \cite{li2023seegera}, and AUGMAE \cite{wang2024rethinking}.

\section{The IMPA-HGAE Approach}

In this section, the model IMPA-HGAE will be introduced, a self-supervised heterogeneous graph neural network based on intra-meta-path enhancement. Specifically, the model employs the following strategies:

\subsection{Meta-Path and Feature Matrix Masking with Reconstruction}

To more effectively leverage long-range information in graphs, IMPA-HGAE employs a dual reconstruction strategy, where both meta-paths and node feature matrices are masked and subsequently reconstructed as learning objectives.Moreover,
IMPA-HGAE employs a more comprehensive masking strategy: in addition to randomly selecting certain nodes for complete feature masking, it also randomly masks partial attribute features of the remaining nodes.

Although mainline methods exhibit good performance on their respective reconstruction targets, they fail to effectively capture long-range information in graphs.

\begin{figure}[!ht]
    \centering
    \includegraphics[width=1\textwidth]{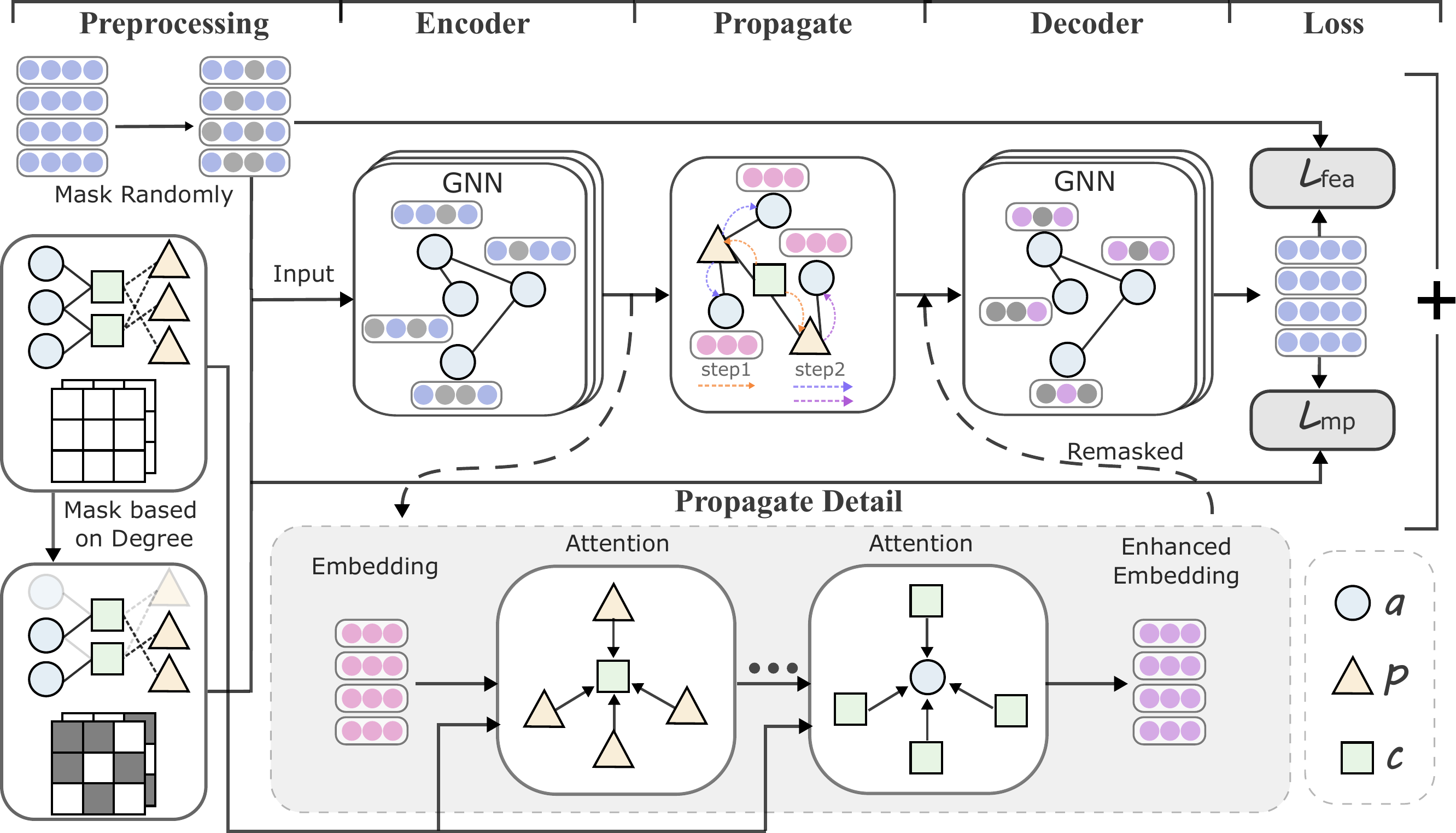}
    \caption{The overall framework of IMPA-HGAE comprises four key modules: data preprocessing, encoding, propagation, and decoding. Initially, the model performs random masking on the feature matrix while applying a degree-based masking strategy to the adjacency matrix. The preprocessed graph data is then encoded via a Graph Neural Network (GNN) to generate node embeddings. During the propagation phase, an attention mechanism is employed to aggregate intra-metapath node information for enhancing the embedding representations. Ultimately, the model simultaneously reconstructs both the adjacency matrix and feature matrix, with joint optimization of their respective reconstruction losses for model training.}
    \label{fig:example2}
\end{figure}

To address this issue, IMPA-HGAE focuses specifically on meta-paths. Particularly, for the input graph, IMPA-HGAE first masks the feature matrices and meta-paths, namely, masks the feature matrices \(\mathcal{X}\) and adjacency matrices \( {\mathcal{A}}^{\Phi} \)of meta-paths. Subsequently, the node embeddings are obtained through the encoder. 

Furthermore, this paper systematically investigates how different meta-path masking strategies affect model performance by introducing distinct noise patterns. Three masking approaches are as follows: (1) random masking, (2) degree-based masking, and (3) attention-score-based masking along meta-paths.

The random masking procedure selects a subset of edges for masking, where the number of masked edges is determined by \( mask\_rate \) × \( |\mathcal{E}| \) , with \( |\mathcal{E}| \) representing the total edge count in the graph. 

For degree-based masking, since the graph is directed, this paper defines the degree-based weight of an edge as the average of its source node's out-degree and destination node's in-degree:
\begin{equation}
 {DW} = Scale( \frac{{DW}_{\text{src}} + {DW}_{\text{dst}}}{2} )
\end{equation}

\begin{equation}
P\left(x_{1}, x_{2}, \ldots, x_{k}\right)={DW}\left[x_{1}\right] \times \frac{DW\left[x_{2}\right]}{1-{DW}\left[x_{1}\right]} \times \cdots \times \frac{{DW}\left[x_{k}\right]}{1-\sum_{j=1}^{k-1} {DW}\left[x_{j}\right]}
\end{equation}

 As mentioned above, the edge sampling is performed using a multinomial distribution based on DW (the tensor recording edge weights), where the sampling probabilities are progressively normalized conditional probabilities. The method prefers to mask edges with higher degree-based weight.

The attention-score-based masking utilizes the edge attention coefficients embedded in the GATConv layers of the HAN encoder. It should be noted that these attention coefficients are node-oriented. 

\begin{figure}[!ht]
    \centering
    \includegraphics[width=0.6\textwidth]{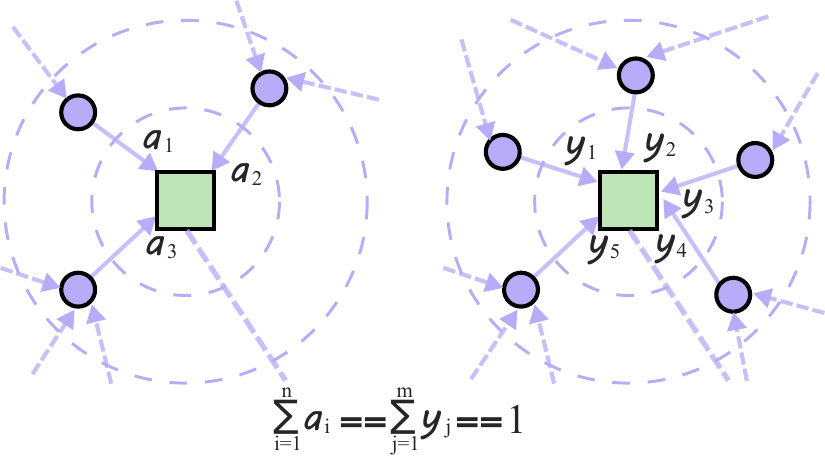}
    \caption{Edge Attention Visualization Diagram}
    \label{fig:example3}
\end{figure}

As illustrated in the Figure \ref{fig:example3} above, the edge attention coefficients are normalized values that depend on the total number of edges pointing to the destination node. To perform the sampling, this approach first applies an inverse softmax transformation with the constant C  to recover the pre-normalized attention scores:

\begin{equation}
z_{i}=\ln \left(p_{i}\right)+C
\end{equation}

These scores are then used for weighted sampling without replacement from a multinomial distribution, following the same procedure as degree-based masking approach.

\subsection{Subgraph Extraction and Information Propagating}
Although meta-paths can reveal deep semantic information in graph structures, most existing neural network models focus only on the path endpoints (target nodes \( i\in tar \), where \( tar \) denotes the target node set), while ignoring non-target nodes \( i\in Ntar \). This approach can lead to information loss and focuses only on a specific type of nodes, neglecting other node types in the graph. For example, in academic networks (authors/papers/venues/keywords), common meta-paths include APA, APVPA, and APTPA. Traditionally, these paths are typically used only to analyze the relationships between authors, often overlooking the associations between papers and authors. This oversight hinders the comprehensive integration of proximity information.

To address this issue and make full use of local proximity information while capturing long-range information, a method involving subgraph extraction and information propagating is proposed. IMPA-HGAE examines feature information from both target and non-target nodes, thereby considering both the structural and feature information of the graph.

First of all, IMPA-HGAE passes the masked feature matrix of the target nodes \( \tilde {\mathcal{X}}^{tar} \) and the masked meta-path adjacency matrix \( \tilde{\mathcal{A}}^{\Phi} \) through the encoder \(f_{E}\) to obtain the feature embeddings of the target nodes \( \mathcal{H}_{\Phi}^{tar} \). For each non-target node, IMPA-HGAE uses a multilayer perceptron (MLP), to map the node embeddings \( \mathcal{X}^{other} \) to the same dimension as the target node embeddings in order to perform subsequent concatenation operations, resulting in the feature embeddings of the non-target nodes.

\begin{equation}
\mathcal{H}^{o}=\sigma^{(2)}\left(W^{(2)}\left(\sigma^{(1)}\left(W^{(1)} \mathcal{X}^{o}+b^{(1)}\right)\right)+b^{(2)}\right), {o}\in \mathcal{T}_{v} \land {o} \neq {tar}
\end{equation}

IMPA-HGAE first extracts multiple adjacent subgraphs \( G_{p,q} \) (where \( p,q \in T_{v}\)) from the original graph data through subgraph sampling. Leveraging meta-path prior knowledge, it then performs subgraph-wise attention-based message passing (see the Figure \ref{fig:example1}), where information propagates from intermediate nodes to terminal nodes. As detailed in the Figure \ref{fig:example2}, this is implemented via stacked generalized GATConv layers - crucially differing from homogeneous GAT architectures by preventing node-level overfitting despite layer stacking.

\begin{equation}
e_{ij}^{\Phi} = \text{att}_{\text{node}}(\mathbf{h}_i, \mathbf{h}_j; e) , e \in \mathcal{T}_{\epsilon}
\end{equation}

Here \( \text{att}_{\text{node}} \) denotes the deep neural network which performs the node-level attention.

\begin{equation}
\alpha_{i j}^{\Phi}=\operatorname{softmax}_{j}\left(e_{i j}^{\Phi}\right)=\frac{\exp \left(\sigma\left(\mathbf{a}_{\Phi}^{\mathrm{T}} \cdot\left[\mathbf{h}_{i} \| \mathbf{h}_{j}\right]\right)\right)}{\sum_{k \in \mathcal{N}_{i}^{\Phi}} \exp \left(\sigma\left(\mathbf{a}_{\Phi}^{\mathrm{T}} \cdot\left[\mathbf{h}_{i} \| \mathbf{h}_{k}\right]\right)\right)},
\end{equation}

Where \( a^{\Phi} \) is the matrix of learnable parameters,\(\mathbf{h}_{i}\) denotes the embedded representation of node i, which is part of \(\mathcal{H}^{\mathcal{T}_{v}(i)}\). \( \mathcal{N}_{i}^{\Phi} \) is the set of neighboring nodes of type \( \Phi \) for node \( i \). The \( \alpha_{ij}^{\Phi} \) is the attention score of \( i \) on \( j \), \( a_{ij}^{\Phi} \) is an asymmetric matrix of attention coefficients indicating that they have different contributions to each other.

\begin{equation}
\mathbf{z}_{i}^{\Phi}=\sigma\left(\sum_{j \in \mathcal{N}_{i}^{\Phi}} \alpha_{i j}^{\Phi} \cdot \mathbf{h}_{j}\right)
\end{equation}

\(\mathbf{z}_{i}^{\Phi}\) is the embedding obtained by the information transfer of the node  \( i \) on the edge of type \({\Phi}\). The final embeddings obtained through multiple layers of attention \( \mathcal{H}^{tar'} \) are processed through the decoder \( f_{D} \) to generate reconstructed outputs \( \mathcal{H}_{D}^{tar} \), which are then used to compute the reconstruction loss.The loss for the reconstructed feature matrix is defined as follows:

\begin{equation}
\mathcal{L}_{\mathrm{feat}}=\frac{1}{|{\mathcal{V}}|} \sum_{tar \in {\mathcal{V}}}\left(1-\frac{{X}_{tar} \cdot H_{D}^{tar}}{\left\|{X}_{tar}\right\| \times\left\|H_{D}^{tar}\right\|}\right)^{\gamma_{1}}
\end{equation}

\begin{equation}
\mathcal{A}^{\phi'} = \delta((\mathcal{H}_D^{tar})^T \cdot (\mathcal{H}_D^{tar}))
\end{equation}

The loss function for the reconstructed meta-path is defined as:

\begin{equation}
\mathcal{L}^\phi=\frac1{|\mathcal{A}^\phi|}\sum_{v\in\mathcal{V}}(1-\frac{\mathcal{A}_v^\phi\cdot\mathcal{A}_v^{\phi\prime}}{\|\mathcal{A}_v^\phi\|\times\|\mathcal{A}_v^{\phi\prime}\|})^{\gamma_{2}}
\end{equation}

\begin{equation}
\mathcal{L}_{\mathrm{mp}}=\sum_{\phi \in \Phi} \alpha^{\phi} \cdot \mathcal{L}^{\phi}
\end{equation}

Among them, \( \alpha^{\phi} \) in HAN denotes the attention coefficients computed for each meta-path,\( \gamma_{1} \) and \( \gamma_{2} \) are the scaling factor.

\section{Experiments}

In this section, this paper demonstrates the effectiveness of IMPA-HGAE by addressing the following issues:

\begin{enumerate}
\item Whether the generative model proposed in this study can handle the downstream task of node classification.

\item Whether the information enhancement method within the meta-path proposed in this study is effective.

\end{enumerate}

\begin{table}[ht!]
\centering
\tiny
\sisetup{
  table-format=2.2,
  table-number-alignment=center,
  table-space-text-post=\textsuperscript{a},
}
\setlength{\tabcolsep}{3pt}
\begin{tabular}{@{}
  >{\centering\arraybackslash}p{0.8cm}
  >{\centering\arraybackslash}p{0.5cm}
  >{\centering\arraybackslash}p{0.2cm}
  >{\centering\arraybackslash}p{1.07cm}
  >{\centering\arraybackslash}p{1.07cm}
  >{\centering\arraybackslash}p{1.07cm}
  >{\centering\arraybackslash}p{1.07cm}
  >{\centering\arraybackslash}p{1.07cm}
  >{\centering\arraybackslash}p{1.07cm}
  >{\centering\arraybackslash}p{1.07cm}
  >{\centering\arraybackslash}p{1.2cm}
  @{}}
\toprule
{Dataset} & {Metric} & {Split} & {HetGNN} & {HAN} & {DMGI} & {HeCo} & {GraphMAE} & {HGMAE} & {DiffGraph} & {IMPA-HGAE} \\
\midrule
\multirow{9}{*}{DBLP} & \multirow{3}{*}{Mi-F1} & 20 & 90.11±1.0 & 90.16±0.9 & 90.78±0.3 & 91.97±0.2 & 89.31±0.7 & \uline{92.71±0.5} & \textbf{92.90±0.4} & 92.09±0.2 \\
 & & 40 & 89.03±0.7 & 89.47±0.9 & 89.92±0.4 & 90.76±0.3 & 87.80±0.5 & \uline{92.43±0.3} & 92.01±0.5 & \textbf{92.68±0.1} \\
 & & 60 & 90.43±0.6 & 90.34±0.8 & 90.66±0.5 & 91.59±0.2 & 89.82±0.4 & \uline{93.05±0.3} & 92.65±0.5 & \textbf{93.20±0.2} \\
\cmidrule(lr){2-11}
 & \multirow{3}{*}{Ma-F1} & 20 & 89.51±1.1 & 89.31±0.9 & 89.94±0.4 & 91.28±0.2 & 87.94±0.7 & \textbf{92.28±0.5} & \uline{92.01±0.2} & 91.64±0.2 \\
 & & 40 & 88.61±0.8 & 88.87±1.0 & 89.25±0.4 & 90.34±0.3 & 86.85±0.7 & \uline{92.12±0.3} & 91.90±0.3 & \textbf{92.39±0.2} \\
 & & 60 & 89.56±0.5 & 89.20±0.8 & 89.46±0.6 & 90.64±0.3 & 88.07±0.6 & \uline{92.33±0.3} & 92.13±0.3 & \textbf{92.35±0.2} \\
\cmidrule(lr){2-11}
 & \multirow{3}{*}{AUC} & 20 & 97.96±0.4 & 98.07±0.6 & 97.75±0.3 & 98.32±0.1 & 92.23±3.0 & 98.90±0.1 & \uline{98.92±0.5} & \textbf{99.12±0.1} \\
 & & 40 & 97.70±0.3 & 97.48±0.6 & 97.23±0.2 & 98.06±0.1 & 91.76±2.5 & \uline{98.55±0.1} & 98.42±0.1 & \textbf{98.63±0.1} \\
 & & 60 & 97.97±0.2 & 97.96±0.5 & 97.72±0.4 & 98.59±0.1 & 91.63±2.5 & 98.89±0.1 & \textbf{99.15±0.3} & \uline{98.95±0.1} \\
\midrule
\multirow{9}{*}{Freebase} & \multirow{3}{*}{Mi-F1} & 20 & 56.85±0.9 & 57.24±3.2 & 58.26±0.9 & 61.72±0.6 & 64.88±1.8 & \uline{65.15±1.3} & 64.55±1.9 & \textbf{65.43±1.5} \\
 & & 40 & 53.96±1.1 & 63.74±2.7 & 54.28±1.6 & 64.03±0.7 & 62.34±1.0 & \textbf{67.23±0.8} & 65.07±1.2 & \uline{66.08±0.9} \\
 & & 60 & 56.84±0.7 & 61.06±2.0 & 56.69±1.2 & 63.61±1.6 & 59.48±6.2 & \uline{67.44±1.2} & 66.32±1.9 & \textbf{67.51±1.5} \\
\cmidrule(lr){2-11}
 & \multirow{3}{*}{Ma-F1} & 20 & 52.72±1.0 & 53.16±2.8 & 55.79±0.9 & 59.23±0.7 & 59.04±1.0 & \uline{62.06±1.0} & \textbf{62.93±1.1} & 62.04±0.8 \\
 & & 40 & 48.57±0.5 & 59.63±2.3 & 49.88±1.9 & 61.19±0.6 & 56.40±1.1 & 64.24±0.9 & \textbf{64.87±0.8} & \uline{64.54±0.5} \\
 & & 60 & 52.37±0.8 & 56.77±1.7 & 52.10±0.7 & 60.13±1.3 & 51.73±2.3 & \textbf{63.84±1.0} & \uline{63.21±1.4} & 63.16±1.0 \\
\cmidrule(lr){2-11}
 & \multirow{3}{*}{AUC} & 20 & 70.84±0.7 & 73.26±2.1 & 73.19±1.2 & 76.22±0.8 & 72.60±0.2 & \uline{78.36±1.1} & 77.72±0.9 & \textbf{78.47±1.1} \\
 & & 40 & 69.48±0.2 & 77.74±1.2 & 70.77±1.6 & 78.44±0.5 & 72.44±1.6 & 79.69±0.7 & \uline{79.79±0.8} & \textbf{79.88±0.7} \\
 & & 60 & 71.01±0.5 & 75.69±1.5 & 73.17±1.4 & 78.04±0.4 & 70.66±1.6 & \uline{79.11±1.3} & \textbf{79.12±1.3} & 78.06±1.2 \\
\midrule
\multirow{9}{*}{ACM} & \multirow{3}{*}{Mi-F1} & 20 & 71.89±1.1 & 85.11±2.2 & 87.60±0.8 & 88.13±0.8 & 82.48±1.9 & 90.24±0.5 & \uline{90.42±1.0} & \textbf{90.46±0.9} \\
 & & 40 & 74.46±0.8 & 87.21±1.2 & 86.02±0.9 & 87.45±0.5 & 82.93±1.1 & 90.18±0.6 & \textbf{90.95±0.7} & \uline{90.89±0.4} \\
 & & 60 & 76.08±0.7 & 88.10±1.2 & 87.82±0.5 & 88.71±0.5 & 80.77±1.1 & \uline{91.04±0.4} & 90.73±0.6 & \textbf{91.24±0.4} \\
\cmidrule(lr){2-11}
 & \multirow{3}{*}{Ma-F1} & 20 & 72.11±0.9 & 85.66±2.1 & 87.86±0.2 & 88.56±0.8 & 82.26±1.5 & 90.66±0.4 & \uline{90.81±0.9} & \textbf{91.05±0.7} \\
 & & 40 & 72.02±0.4 & 87.47±1.1 & 86.23±0.8 & 87.61±0.5 & 82.00±1.1 & 89.15±0.6 & \uline{89.96±0.9} & \textbf{91.02±0.4} \\
 & & 60 & 74.33±0.6 & 88.41±1.1 & 87.97±0.4 & 89.04±0.5 & 80.29±1.0 & 90.59±0.4 & \textbf{91.73±0.7} & \uline{91.39±0.2} \\
\cmidrule(lr){2-11}
 & \multirow{3}{*}{AUC} & 20 & 84.36±1.0 & 93.47±1.5 & 96.72±0.3 & 96.49±0.3 & 92.09±0.5 & \uline{97.69±0.1} & 97.56±0.3 & \textbf{97.70±0.4} \\
 & & 40 & 85.01±0.6 & 94.84±0.9 & 96.35±0.3 & 96.40±0.4 & 92.65±0.5 & 97.52±0.1 & \uline{97.72±0.3} & \textbf{98.08±0.1} \\
 & & 60 & 87.64±0.7 & 94.68±1.4 & 96.79±0.2 & 96.55±0.3 & 91.49±0.6 & \uline{97.87±0.1} & 97.66±0.2 & \textbf{98.13±0.1} \\
\midrule
\multirow{9}{*}{Aminer} & \multirow{3}{*}{Mi-F1} & 20 & 61.49±2.5 & 68.86±4.6 & 63.93±3.3 & 78.81±1.3 & 68.21±0.3 & \uline{80.30±0.7} & 79.99±0.7 & \textbf{80.64±1.1} \\
 & & 40 & 68.47±2.2 & 76.89±1.6 & 63.60±2.5 & 80.53±0.7 & 74.23±0.2 & \uline{82.35±1.0} & 81.36±1.1 & \textbf{83.42±1.1} \\
 & & 60 & 65.61±2.2 & 74.73±1.4 & 62.51±2.6 & \uline{82.46±1.4} & 72.28±0.2 & 81.69±0.6 & 81.10±1.2 & \textbf{83.27±0.6} \\
\cmidrule(lr){2-11}
 & \multirow{3}{*}{Ma-F1} & 20 & 50.06±0.9 & 56.07±3.2 & 59.50±2.1 & 71.38±1.1 & 62.64±0.2 & \uline{72.28±0.6} & 71.40±0.9 & \textbf{72.91±1.1} \\
 & & 40 & 58.97±0.9 & 63.85±1.5 & 61.92±2.1 & 73.75±0.5 & 68.17±0.2 & \uline{75.27±1.0} & 75.37±1.0 & \textbf{76.27±1.1} \\
 & & 60 & 57.34±1.4 & 62.02±1.2 & 61.15±2.5 & \uline{75.80±1.8} & 68.21±0.2 & 74.67±0.6 & 75.32±0.6 & \textbf{76.28±0.9} \\
\cmidrule(lr){2-11}
 & \multirow{3}{*}{AUC} & 20 & 77.96±1.4 & 78.92±2.3 & 85.34±0.9 & 90.82±0.6 & 86.29±4.1 & \textbf{93.22±0.6} & 90.12±1.1 & \uline{92.40±0.6} \\
 & & 40 & 83.14±1.6 & 80.72±2.1 & 88.02±1.3 & 92.11±0.6 & 89.98±0.0 & \uline{94.68±0.4} & 94.34±0.4 & \textbf{94.75±0.3} \\
 & & 60 & 84.77±0.9 & 80.39±1.5 & 86.20±1.7 & 92.40±0.7 & 88.32±0.0 & \uline{94.59±0.3} & 94.12±0.7 & \textbf{94.63±0.2} \\
\bottomrule
\end{tabular}
\vspace{5pt} 
\caption{Performance comparison of node classification tasks. The best results are highlighted in bold, while the second-best results are marked with underlines.}
\label{tab:full_results}
\end{table}

\subsection{Node Classification}

\textbf{Datasets and Baselines.} Four real-world data sets, including DBLP, Freebase, ACM, and AMiner, are used to evaluate the proposed model. Comparisons are made with ten baseline methods, including unsupervised homogeneous graph methods GraphMAE \cite{hou2022graphmae}, unsupervised/semi-supervised heterogeneous graph methods HetGNN \cite{zhang2019heterogeneous}, DMGI \cite{park2020unsupervised}, HeCo \cite{wang2021self}, and HAN \cite{wang2019heterogeneous}. Additionally, this paper also compares with the latest model for handling heterogeneous graphs, HGMAE and DiffGraph \cite{li2025lncs_preprint}.

The performance of IMPA-HGAE was evaluated on four datasets: DBLP, Freebase, ACM, and AMiner. For each class, 20\%, 40\%, and 60\% of the nodes were allocated as training data, with the remaining nodes serving as the test and validation sets. The evaluation metrics included Micro-F1 (Mi-F1), Macro-F1 (Ma-F1), and AUC.

\begin{figure}[h!btp]
    \centering
    \includegraphics[width=0.7\textwidth]{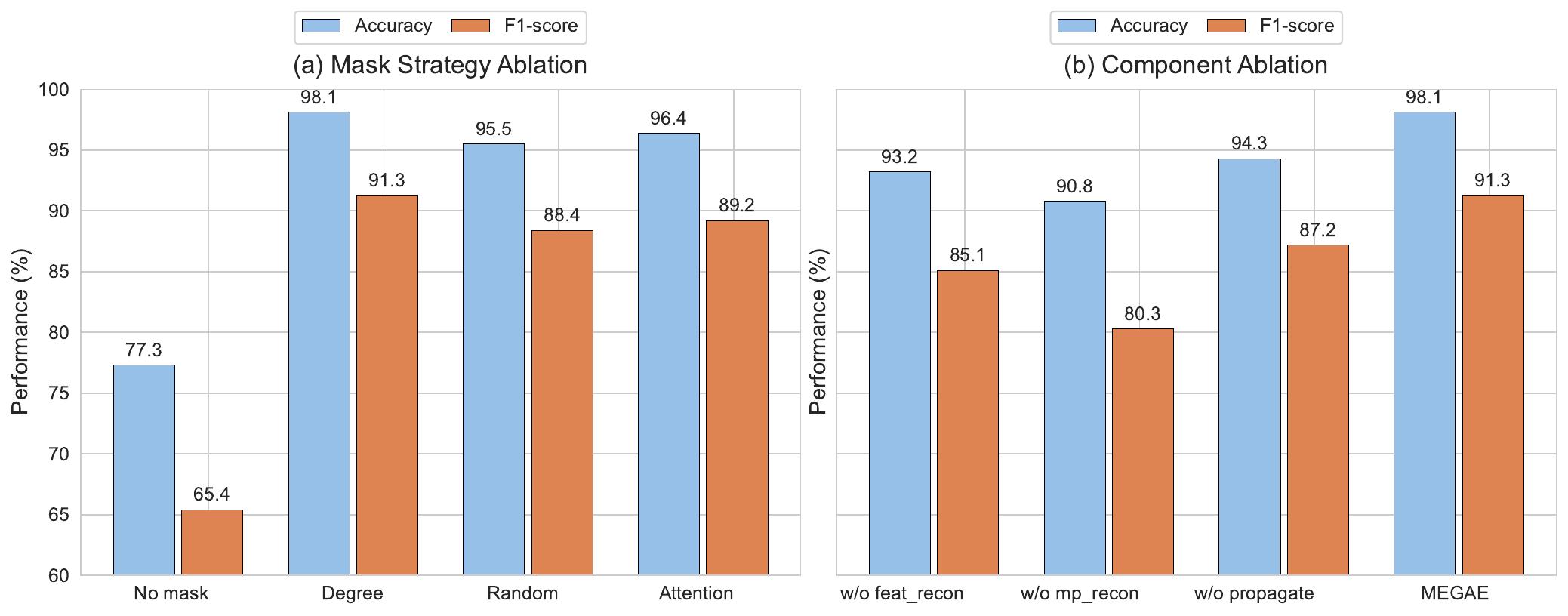}
    \caption{Ablation Study on Masking Strategies and Model Components on ACM Dataset}
    \label{fig:example4}
\end{figure}

\subsection{Ablation Study}

Through ablation experiments (see the Figure \ref{fig:example3}), it was found that applying IMPA-HGAE improves model accuracy and F1-score on the ACM dataset, and the same applies to other datasets. 

From the left figure of Figure \ref{fig:example4}, it is evident that implementing the masking operation on the ACM dataset significantly boosts the performance of the node classification task, as reflected by improvements in two key metrics: accuracy and F1-score. Among various masking strategies, the degree-based masking mechanism exhibits the most substantial performance enhancement, followed by attention-score-based masking, while random masking demonstrates relatively weaker performance. 

The experimental results presented in the right figure further indicate that independently reconstructing either the feature matrix or meta-paths performs less effectively compared to the IMPA-HGAE model. This comparison strongly underscores the significant advantage of the IMPA-HGAE design strategy, which simultaneously considers the feature matrix and meta-paths as reconstruction targets. Additionally, incorporating neighboring node enhancement significantly improves model performance, thereby validating the effectiveness of enhancing internal node information within meta-paths. Both conclusions have been verified across multiple datasets, demonstrating their generalizability. 

\begin{figure}[htbp]
    \centering
    \includegraphics[width=1\textwidth]{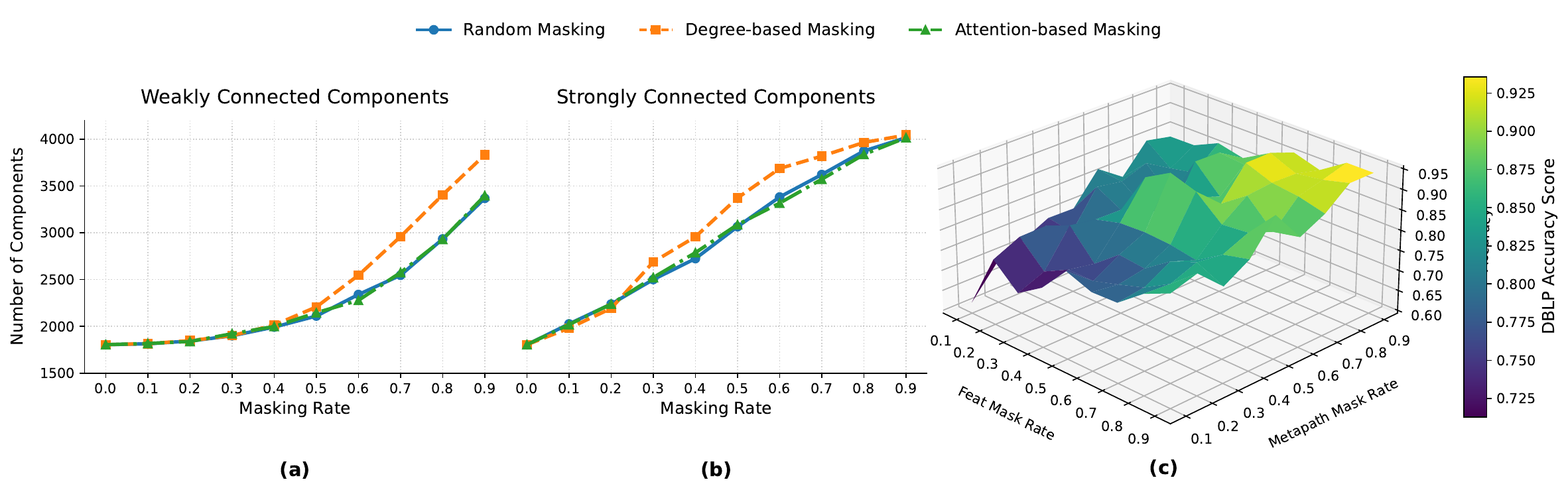}
    \caption{Mask Analysis Diagram: (a)(b) Connected component trends under various masking ratios and methods; (c) 3D surface comparison between feature masking and meta-path masking.}
    \label{fig:example5}
\end{figure}
\vspace{-0.5cm}

\subsection{Mask Policy Analysis}

According to the left figure in Figure \ref{fig:example4}, the degree-based masking strategy achieves the best performance, whereas others exhibit slightly inferior results. A deeper investigation is warranted here. To evaluate the loss of structural information in the graph, this article assess the number of connected components in the masked graph. It is observed (as depicted in Figure \ref{fig:example5}) that degree-based masking generates a higher number of connected components at the same masking rate, while the number of connected components generated by random masking and attention-score-based masking tends to be consistent. 

Combining the experimental results from Figure \ref{fig:example4}, this hypothesize that degree-based masking preferentially masks edge-dense regions of the graph, thereby producing more uniform graph data. In contrast, random masking and attention-score-based masking fail to achieve this effect. Furthermore, the translation invariance property of the softmax function implies that restoring attention masking using the inverse softmax function may only preserve relative magnitudes, potentially affecting the global accuracy of attention coefficients. Due to space constraints, this article expects further in-depth research into this area. 

\section{Conclusions and Future Scope}

\begin{figure}[htbp]
    \centering
    \includegraphics[width=1\textwidth]{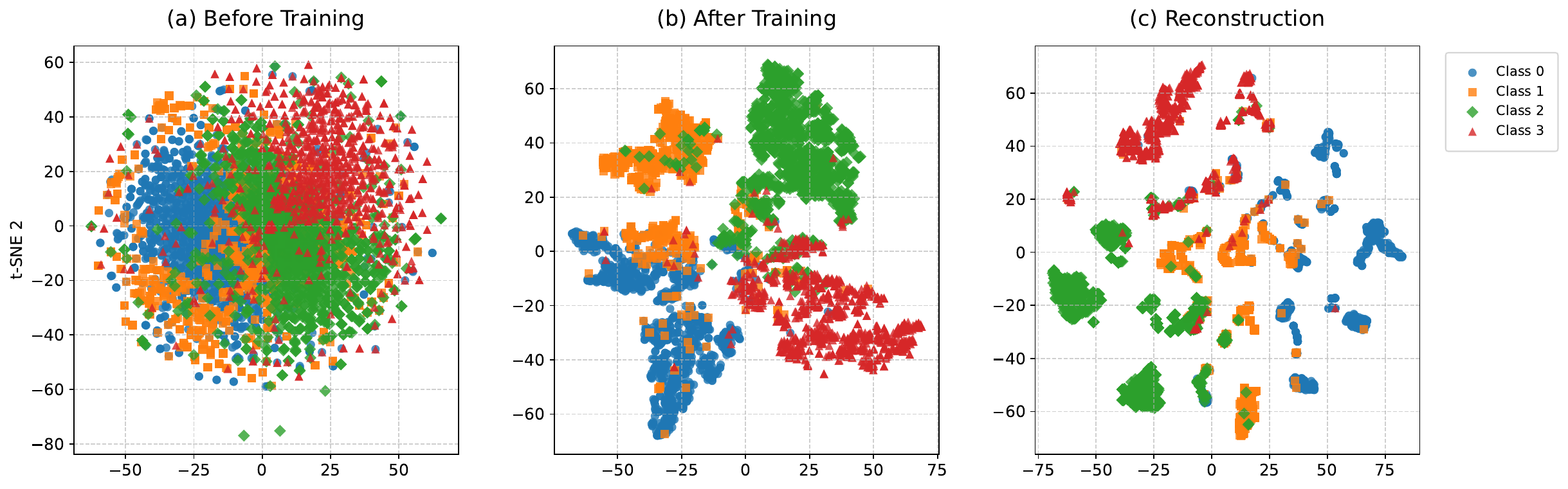}
    \caption{Visualization of DBLP Dataset Distribution via t-SNE Dimensionality Reduction}
    \label{fig:example6}
\end{figure}

This paper proposes a self-supervised heterogeneous graph neural network model, IMPA-HGAE, which achieves meta-path masking and reconstruction as its core objective. Furthermore, it integrates the enhancement method introduced in this study. Experimental results indicate that IMPA-HGAE significantly outperforms various baseline models. Overall, IMPA-HGAE introduces new advancements to self-supervised generative learning and heterogeneous graph neural networks, showcasing the substantial potential of graph generative learning. Figure \ref{fig:example6} presents the t-SNE visualization of data processed by IMPA-HGAE at different stages. It is evident that the obtained embeddings effectively distinguish the original data. Meanwhile, Figure \ref{fig:example6}(c) illustrates the visualization of outputs generated by the decoder. Intuitively, the training objective should converge to the representation depicted in Figure \ref{fig:example6}(a), but the results are not entirely satisfactory. Consequently, this study suggests that future generative self-supervised models should either enhance the decoder's capability or design auxiliary tasks that better facilitate the decoder's performance.

%
%
%
%

\end{document}